\documentclass[a4paper,conference]{IEEEtran}

\usepackage{times}
\usepackage{soul}
\usepackage{url}
\usepackage[hidelinks]{hyperref}
\usepackage[utf8]{inputenc}
\usepackage[small]{caption}
\usepackage{graphicx}
\usepackage{amsmath}
\usepackage{amsfonts}
\usepackage{amsthm}
\usepackage{booktabs}
\usepackage{algorithm}
\usepackage{algorithmic}
\usepackage{verbatim}
\usepackage{multirow}
\usepackage{diagbox}
\usepackage{booktabs}
\ifCLASSINFOpdf
\else
\fi

\graphicspath{{D:/PhD/Paper/IJCAI2020/figures/}}

\begin{document}
\title{Interpreting the Latent Space of GANs via Correlation Analysis for Controllable Concept Manipulation}



\author{\IEEEauthorblockN{Ziqiang Li$^*$, Rentuo Tao$^*$, Hongjing Niu, Bin Li}
	\IEEEauthorblockA{CAS Key Laboratory of Technology in Geo-spatial Information Processing and Application Systems\\University of Science and Technology of China, Anhui, China\\\{iceli, trtmelon,sasori\}@mail.ustc.edu.cn, binli@ustc.edu.cn}}


%


\maketitle


\begin{abstract}
	Generative adversarial nets (GANs) have been successfully applied in many fields like image generation, inpainting, super-resolution, and drug discovery, etc. By now, the inner process of GANs is far from being understood. To get a deeper insight into the intrinsic mechanism of GANs, in this paper, a method for interpreting the latent space of GANs by analyzing the correlation between latent variables and the corresponding semantic contents in generated images is proposed. Unlike previous methods that focus on dissecting models via feature visualization, the emphasis of this work is put on the variables in latent space, i.e. how the latent variables affect the quantitative analysis of generated results. Given a pre-trained GAN model with weights fixed, the latent variables are intervened to analyze their effect on the semantic content in generated images. A set of controlling latent variables can be derived for specific content generation, and the controllable semantic content manipulation is achieved. The proposed method is testified on the datasets Fashion-MNIST and UT Zappos50K, experiment results show its effectiveness.
\end{abstract}
\renewcommand{\thefootnote}{}
\footnote{$^*$Equal Contribution}
\footnote{Accepted by ICPR2020}
\begin{figure}[ht]
	\includegraphics[width=0.4\textwidth]{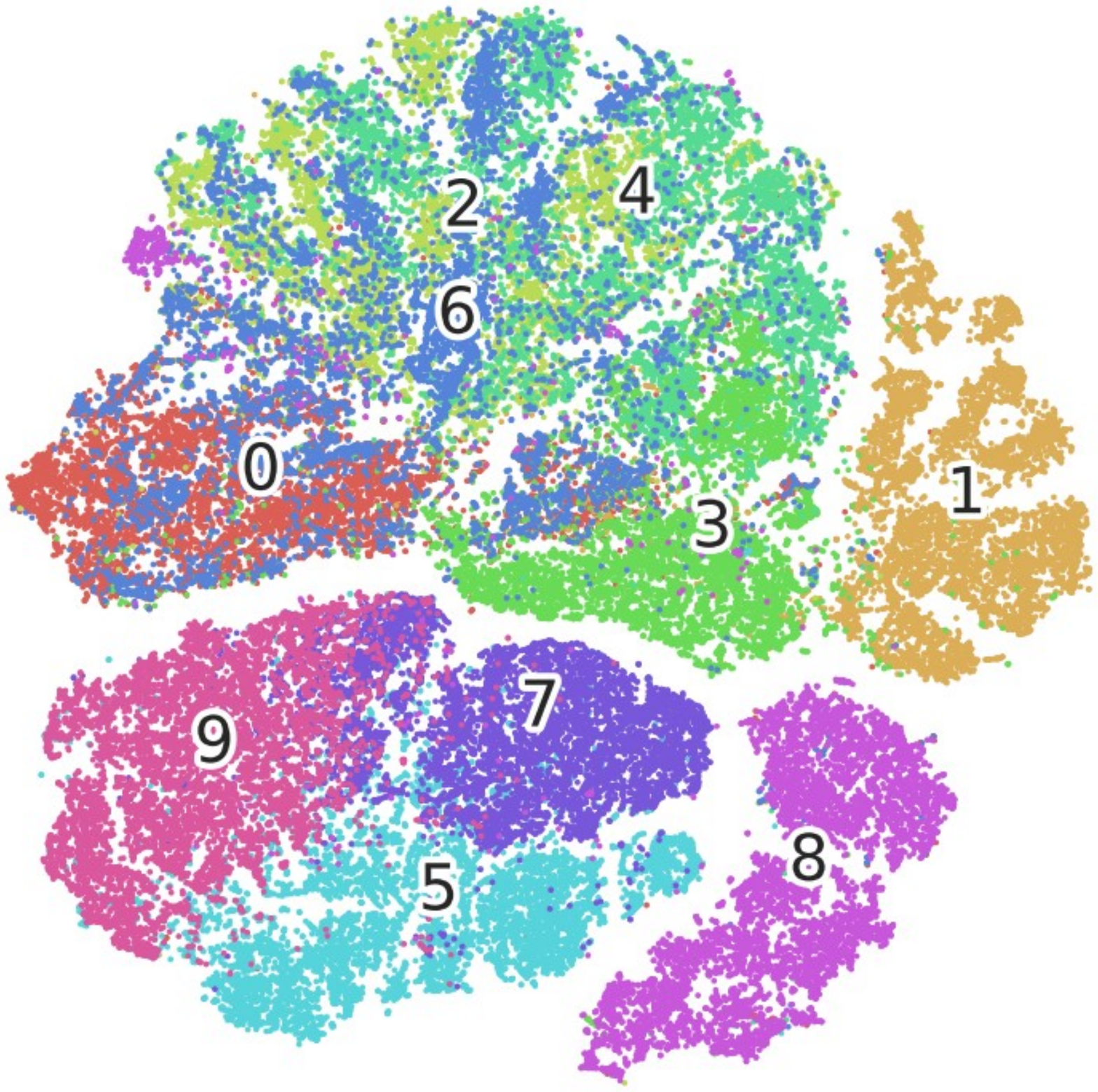}
	\centering
	\caption{t-SNE analysis on latent representations of Fashion-MNIST dataset. Points in different color are 2D features of latent representations belong to different classes.}
	\label{fig:tsne}
\end{figure}
\section{Introduction}
GANs\cite{goodfellow2014generative} have achieved great success in many tasks like image generation\cite{brock2018large}\cite{karras2019style}, super resolution\cite{ledig2017photo} and image translation\cite{lample2017fader}\cite{choi2018stargan}. However, due to the great number of parameters and a complex cascade of nonlinear activations, we can hardly understand the intrinsic logic of GANs and the quantitative relationship between input values of variables in latent space and semantic content in output generated images. Moreover, GANs still suffer some severe issues, such as unstable training, mode drop and mode collapse, etc. Many works toward mitigating these issues have been proposed: designing new architectures\cite{karras2017progressive}\cite{brock2018large}, new loss functions\cite{arjovsky2017wasserstein}\cite{gulrajani2017improved} or new training methologies\cite{karras2017progressive}\cite{karras2019style}. Compared to the large number of works that aim at improving train stability and generation quality, few works have been done to explore the intrinsic working mechanism of GANs, i.e. the interpretability of GANs. This motivated the work of this paper.

Recently, along with the success of deep learning, interpreting deep neural models have become a hot topic in research. The methods can be generally divided into three categories according to the corresponding interpreting results: visualizing pre-trained models\cite{zeiler2014visualizing}\cite{mahendran2015understanding}, diagnose pre-trained models\cite{ribeiro2016should}\cite{lakkaraju2017identifying} and construct interpretable models\cite{zhang2019interpreting}\cite{zhang2018interpretable}. The first kind of methods use activation maximization or de-convolution techniques to visualize neurons or derive an input that maximizes semantic outputs; the second kind of methods diagnose pre-trained deep models by adding understandable disturbance to the inputs or quantify the importance of features; the last kind of methods impose constraints on neurons or reconstruct deep models to make deep models more interpretable. As for GANs, researchers proposed various methods to make them more interpretable and controllable, such as concatenating latent code with conditional information\cite{mirza2014conditional}\cite{goetschalckx2019ganalyze}, interpolating in latent manifolds\cite{shen2019interpreting}\cite{jahanian2019steerability} and latent disentanglement\cite{lample2017fader}\cite{donahue2017semantically}. Bau et.al\cite{bau2018gan} proposed to quantify the effect or importance of different layer nodes on content changes by utilizing a pre-trained segmentation model to calculate IoU scores on generated image variations. However, previous methods did not give any quantitative evaluation of contribution each variable in latent space provides for generating specific semantic contents.

For the interpretability of GANs, due to the black-box property of deep neural models, hardly can we understand how the latent variables affect the generation process. To investigate if the variables in latent space contain the necessary information to distinguish different semantic contents in generated images, we use t-SNE\cite{maaten2008visualizing} to analyze the latent representations of Fashion-MNIST samples and find that the latent representations of samples from different classes can be well-separated, as shown in Figure \ref{fig:tsne}. This indicates that for samples from the same one class, there may be exist closely-related latent variables that made their latent representations distinguishable. This motivated us to quantify the importance of different latent dimensions for specific concept generation. We propose to analyze the correlation between the latent inputs and the corresponding generated outputs by utilizing a pre-trained classifier to provide quantitative evaluations on the semantic content changes in generated images, then to interpret the meaning of variables in the latent space of GANs. We also propose two methods for locating high-correlated latent dimensions for specific concept generation or manipulation: one is by sequential intervention and the other is by optimization. Moreover, given a specific class concept, we can also intervene in the derived controlling set of latent variables for controllable content manipulation. Specifically, the main contributions of this paper are:

\begin{itemize}
	\item We first propose to interpret the latent space of GANs by quantifying the correlation between the latent inputs and the generated outputs.
	\item We demonstrate that for generating contents of specific concept, the importance of different latent variables may varies greatly. Moreover, we propose an optimization-based method to find controlling latent variables for specific concept.
	\item The proposed method can fulfill controllable concept manipulation in generated images via controlling variables discovering and intervention.
\end{itemize}

The remaining part of this paper is organized as follows: in section 2, we give a brief introduction to the backgrounds of this work. In section 3, we introduc the proposed method for interpreting the latent space of GANs in detail. The experiment results and implementation details are given in section 4 and finally is the conclusion of this paper.

\begin{figure*}[h]
	\includegraphics[width=0.75\textwidth]{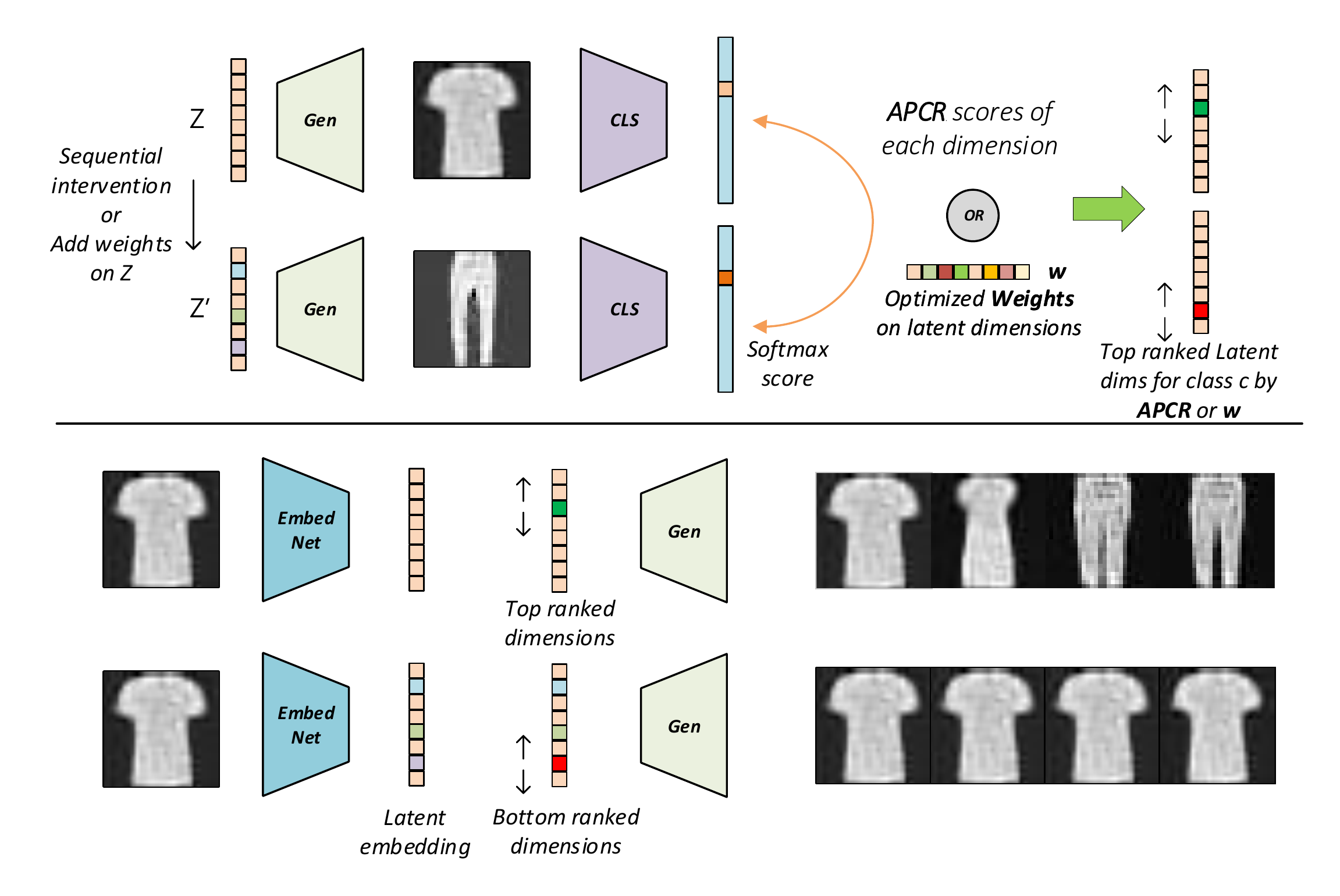}
	\centering
	\caption{The proposed method for analyzing the correlation between latent space and output image space of GANs. Top part illustrate the process of finding high-correlated latent dimensions by sequential intervention or adding weights on latent variables. Bottom part denote the process of latent intervention on top or bottom ranked latent dimensions.}
	\label{fig:CorrelationGAN}
\end{figure*}
\section{Preliminaries}
\subsection{Generative Adversarial Networks}
Generative adversarial nets (GANs) aimed at modeling the train data distribution through an adversarial process. It mainly contains two models that play against each other, a generator G and a discriminator D, G maps noise inputs from latent space to image space and tries to let D classify the generated samples as true, whereas D wants to clearly distinguish real data from generated ones. The adversarial process can be formulated as below:
\begin{align}
\begin{split}
\min_{G}\max_{D}V(D,G)=&\mathbb{E}_{\mathbf{x}\sim p_{data}(\mathbf{x})}[\log D\left( \mathbf{x}\right)]\\
&+\mathbb{E}_{\mathbf{z}\sim p_{z}(\mathbf{z})}[\log \left( 1-D\left( G\left( \mathbf{z}\right) \right)\right) ]
\end{split}
\end{align}
where $V(D,G)$ is the value function of the min-max game, $p_{data}(\mathbf{x})$ and $p_{z}(\mathbf{z})$ represent the distribution of train data and latent noise respectively. In practice, people tend to train G to maximize $\log D(G(\mathbf{z}))$ rather than to minimize $\log (1-D(G(\mathbf{z})))$ for the reason that equation 1 may not provide sufficient gradient for G in the early training stage. In this paper, we choose a pre-trained WGAN\cite{arjovsky2017wasserstein} model for latent space analysis.

\subsection{GAN Dissection}
In \cite{bau2018gan}, the author proposed an analytic framework to visualize and understand GANs at unit-, object- and scene level respectively by identifying interpretable units that are closely related to some object concepts through a segmentation network. Moreover, the authors also quantify the causal effect of interpretable units and offer an empirical method to mitigate the artifacts problem in generation results. The method for characterizing units by dissection can be formulated as below: 
\begin{equation}
IoU_{u,c}\equiv \frac{\mathbb{E}_{\mathbf{z}}\left|\left(\mathbf{r}_{u,\mathbb{P}}^{\uparrow}>t_{u,c}\right)\wedge\mathbf{s}_{c}\left(\mathbf{x}\right)\right|}{\mathbb{E}_{\mathbf{z}}\left|\left(\mathbf{r}_{u,\mathbb{P}}^{\uparrow}>t_{u,c}\right)\vee\mathbf{s}_{c}\left(\mathbf{x}\right)\right|}
\end{equation}
where $u$ denotes the layer node in GANs and $c$ means concepts like classes, $\mathbf{r}_{u,\mathbb{P}}^\uparrow$ denote the feature map of unit $u$, $\wedge$ and $\vee$ represent intersection and union operation respectively, $t_{u,c}$ is a chosen threshold for producing binary mask and $s_{c}\left( \mathbf{x}\right)$ is the segmentation result of the generated image $\mathbf{x}$. This method uses the IoU (Intersection of Union) score to quantify the spatial agreement between a unit's feature map and the segmentation map of a generated image, which can reflect the importance of a unit for generating a specific concept.
\section{Correlation Analysis between Latent and Output Spaces}
In this paper, we propose to interpreting the latent space of GANs by analyzing the correlation between latent dimensions and the corresponding content changes in generated images. A thing that needs to notice is that the proposed method aimed at analyzing a pre-trained GAN model other than training a new one during the correlation analysis. The target was to semantically interpret the latent space and quantify the importance of different latent dimensions, below are the details of the proposed method.
\subsection{Problem Statement}
Given a pre-trained generator $\mathbf{G}\left(\mathbf{z}\right)$, the target was to analyze the latent space and quantify the importance of different latent dimensions for a generation. As for specific semantic concepts like classes or objects, we want to analyze which dimensions influence the contents of the generated results most. The problem can be formulated as below:
\begin{align}
\begin{split}
\mathbf{X}=&\mathbf{G}\left(\mathbf{Z}\right)=\mathbf{G}\left(\left[z_1,z_2,\dots,z_N\right]\right)\\
\Delta z_i \Longrightarrow &\Delta \mathbf{X} \Longrightarrow \Delta \mathbf{C}_l, \quad \textrm{for} \,\, i=1,\dots,N
\end{split}
\end{align}
where $\left[z_1,z_2,...,z_N\right]=\mathbf{z}\in \mathbb{R}^N$ is denoted as the latent variable, $\Delta\mathbf{X}$ and $\Delta \mathbf{C}_l$ ($\mathbf{C}\in \mathbb{R}^L$, $L$ classes) represent the content change and concept(here we choose class labels as different concepts) change in the corresponding generated image. For each latent dimension, we adjust the value of this dimension and observe how much can the intervention influence the generation results. Practically, as will be introduced in the latter part, for most concepts, there may be more than one latent dimension that closely related to that concept, thus we also try to find these controlling dimensions for different concepts. The problem can be simply formulated as:
\begin{equation}
\begin{split}
\alpha_i \in \{0,1\},\quad &i=1,\dots,N\\
\left[\alpha_1 \cdot z_1,\dots,\alpha_N \cdot z_N\right] &\Longleftrightarrow \Delta\mathbf{X} \Longleftrightarrow \Delta \mathbf{C}_l
\end{split}
\end{equation}
where $\alpha_i \in \{0,1\}$ represents whether the corresponding dimension is high-correlated with a given concept $\mathbf{C}_l$ or not. In the next two subsections, we will introduce the proposed method for finding important latent dimension by quantifying the semantic content change and finding controlling dimensions for the given concept through optimization.
\subsection{High-correlated latent dimensions for specific concept generation}
To analyze the correlation between latent space and the output image space, we adopted a pre-trained classifier $\mathbf{Q}\left(\mathbf{x}\right)$ to assign a score on the generated image and its variations. The pre-trained classifier $\mathbf{Q}$ can quantify the content change of the base generated image as a softmax score, which can be used for latent dimension importance analysis. For i-th latent dimension and j-th class
\begin{align}
\mathbf{Z}^k=&\mathbf{Z}+k\cdot \delta\cdot\left[0,\dots,1,\dots,0\right]\quad k\in\left[-m,m\right]\\
\mathbf{X}^k=&\mathbf{G}\left(\mathbf{Z}^k\right)=\mathbf{G}\left(\left[z_1,\dots,z_i+\delta\cdot k,\dots,z_N\right]\right)\\
&\mathbf{S}_{i,j}^k=\mathbf{Q}_j\left(\mathbf{X}^k\right)=\mathbf{Q}_j\left(\mathbf{G}\left(\mathbf{Z}^k\right)\right)
\end{align}
where $\mathbf{Z}^k$, $\mathbf{X}^k$ and $\mathbf{S}_{i,j}^k$ represent the latent variable(intervened the i-th dimension), the generated image and the corresponding classification softmax score of class $j$ while $k$ adjust the value of i-th latent dimension bi-directional k steps(the stepsize was $\delta$). It's easy to find that when $k=0$, $\mathbf{Z}^0$ and $\mathbf{X}^0$ is just the same as $\mathbf{Z}$ and $\mathbf{X}$, i.e., the base reference generated image and latent variable. To measure the i-th latent dimension's effect on the generation of j-th class contents, we use the averaged probability change ratio (APCR) as the quantitative evaluation metric. 
\begin{align}
\begin{split}
{APCR}_{i,j}=&\left\|\frac{\sum_{k=1}^{m}\left(\mathbf{S}_{i,j}^k-\mathbf{S}_{i,j}^{k-1}\right)}{2\cdot m}\right\|_1\\
+&\left\|\frac{\sum_{k=-1}^{-m}\left(\mathbf{S}_{i,j}^k-\mathbf{S}_{i,j}^{k-1}\right)}{2\cdot m}\right\|_1
\end{split}
\end{align}

As can be seen from the above equation, the metric calculates the averaged value changes of the probability that the generated image variations belong to a specified class over the latent dimension changes. Here we use the L-1 norm to calculate the probability changes separately along with two variation directions. Then, for each class, we will get $n$ (dimension of the latent space) APCR scores and find the most correlated dimension for this class by ranking the scores. Similarly, we can also derive $L$ APCR scores for each latent dimension.
\subsection{Controlling set of latent dimensions for controllable concept manipulation}
As mentioned above, for a given specific class, there may have multi latent dimensions that closely related to it, hence we propose to find these important dimensions, which we called controlling latent dimension here. We propose to intervene in all the latent dimensions by assigning different weights to these dimensions and observe the corresponding classification score changes. The optimization objective was to maximize the classification score changes by updating the coefficients on the latent dimensions. The model weights were keep frozen during the optimization process. To derive the controlling dimensions for j-th class, we firstly add differentiated distortions to the latent dimensions by utilizing a weight vector $\mathbf{w}=\left[w_1,\dots,w_N\right]$:
\begin{equation}
\mathbf{Z}^{'}=\mathbf{Z}+\mathbf{w}*\xi=\left[z_1+w_1\cdot \xi,\dots,z_N+w_N\cdot \xi\right]
\end{equation}
where $\xi$ represent an experimental constant for latent intervention, and $w_i\in[-1,1]$ represent the latent intervention coefficients. Positive coefficients denote positive intervention direction and vise versa. For j-th class to be analyzed, we first calculate the probability changes with respect to latent interventions. Then we can derive the optimization objective of the weight vector $\mathbf{w}$:
\begin{align}
\Delta\mathbf{S}_j=\mathbf{S}_j^{'}-\mathbf{S}_j=\mathbf{Q}_j\left(\mathbf{G}\left(\mathbf{Z}^{'}\right)\right)-\mathbf{Q}_j\left(\mathbf{G}\left(\mathbf{Z}\right)\right)\\
\mathbf{w}^*=\arg\max \mathbf{L}_{\mathbf{w}}=\arg\max\left(\left|\Delta\mathbf{S}_j\right|\right)
\end{align}
For easy optimization and make the coefficients on latent dimensions sparse, we add the L-2 norm regularization and rewrite the optimization objective, the optimal $\mathbf{w}$ can be derived through the optimization of equation (12). Finally, we add a threshold $t_j$ on the optimized coefficients $\mathbf{w}$ to filter out uncorrelated dimensions for each class $j$, hence the controlling dimensions can be derived.
\begin{align}
\mathbf{w}^*=\arg\min\left(1-\left|\Delta\mathbf{S}_j\right|+\lambda\cdot\left\|\mathbf{w}\right\|_2\right)\\
\mathbf{w}_{>t_j}\Longleftarrow\left[\left|w_{j,1}^*\right|>t_j,\dots,\left|w_{j,N}^*\right|>t_j\right]
\end{align}

Moreover, for controllable concept manipulation, we can just modify the optimization objective as below and derive the controlling set of latent dimensions for class-to-class translation (class j and class k).
\begin{align}
\begin{split}
\Delta\mathbf{S}_{k\rightarrow j}&=\mathbf{Q}_j\left(\mathbf{G}\left(\mathbf{Z}_k-w*\xi\right)\right)-\mathbf{Q}_j\left(\mathbf{G}\left(\mathbf{Z}_k\right)\right)\\
\Delta\mathbf{S}_{j\rightarrow k}&=\mathbf{Q}_k\left(\mathbf{G}\left(\mathbf{Z}_j+w*\xi\right)\right)-\mathbf{Q}_k\left(\mathbf{G}\left(\mathbf{Z}_j\right)\right)
\end{split}
\end{align}
\begin{align}
\begin{split}
\mathbf{w}^*&=\arg\max \mathbf{L}_{\mathbf{w}}=\arg\max\left(\left|\Delta\mathbf{S}_{k\rightarrow j}\right|+\left|\Delta\mathbf{S}_{j\rightarrow k}\right|\right)\\
&=\arg\min\left(2-\left|\Delta\mathbf{S}_{k\rightarrow j}\right|-\left|\Delta\mathbf{S}_{j\rightarrow k}\right|+\lambda\cdot\left\|\mathbf{w}\right\|_2\right)
\end{split}
\end{align}
Thus for $Z_j$ and $Z_k$ that belong to class $j$ and $k$, we can achieve class j $\leftrightarrow$ class k translation by $Z_j+w*\xi$ and $Z_k-w*\xi$ respectively. The proposed method was depicted in Figure \ref{fig:CorrelationGAN}, where the top part denotes the process of finding high-correlated latent dimensions for the specific concept by sequential intervention and adding weights on latent variables. The bottom part denotes the process of controllable manipulation through latent intervention on top-ranked and bottom-ranked dimensions.

\section{Experiments}
In this section, we will present the experiment results to demonstrate the effectiveness of the proposed method along with some interesting findings. The experiments can be divided into four parts, the first one is designed to find high-correlated latent dimensions for the specific concept by intervening latent dimensions sequentially, the second one solve the same problem as the first but with an optimization-based method, the third experiment is about controllable concept manipulation using controlling latent dimensions and the last one is about class2class image translations.
\subsection{Datasets and Implementation Details}
In this paper, we conducted designed experiments on two datasets: Fashion-MNIST\cite{xiao2017/online} and UT Zappos50K\cite{semjitter}. Fashion-MNIST is a dataset of Zalando's article images—consisting of 60k training samples and 10k testing samples, just the same with the original MNIST dataset. Each example is a 28x28 grayscale image, associated with a label from 10 classes: [\textit{T-shirt, Trouser, Pullover, Dress, Coat, Sandal, Shirt, Sneaker, Bag, Ankle boot}]. UT Zappos50K is a large shoe dataset consisting of 50,025 catalog images collected from Zappos.com. The images are divided into 4 major categories — shoes, sandals, slippers, and boots — followed by functional types and individual brands. As for the implementation, we use a pre-trained WGAN\_GP model for correlation analysis. Moreover, we adopt the LeNet-5 model as the choice for classification network and EmbedNet. The length of latent variables is set to 100.
\subsection{Find High-Correlated Latent Dimensions by Sequential Latent Intervention}
\begin{figure}[h]
	\centering
	\includegraphics[width=0.45\textwidth]{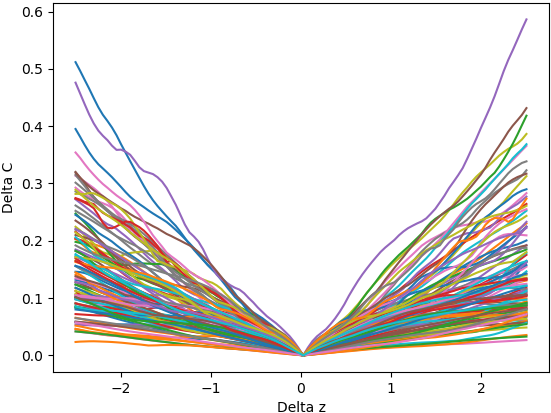}
	\caption{Classification score change with respect to intervention on different latent dimensions. Each color represent a latent dimension.}
	\label{fig:Sequential}
\end{figure}

\begin{figure}[h]
	\centering
	\includegraphics[width=0.45\textwidth]{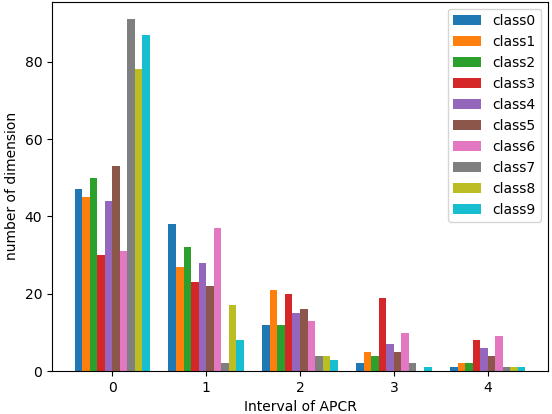}
	\caption{Number distribution of latent dimensions with respect to different APCR value range.}
	\label{fig:APCR}
\end{figure}

As having been introduced in Section 3.2, given a specific class concept, we intervene in the latent dimensions sequentially and get corresponding APCR scores for each latent dimension. Then we rank the latent dimensions for each concept according to the APCR values, thus high-correlated latent dimensions can be derived. Here we use the softmax score to indicate the concept changes in generated images. The softmax score changes w.r.t latent offsets for each latent dimension was shown in Figure \ref{fig:Sequential}, from which we can clearly see that the slope of different curves varies greatly. It indicates that semantic concepts are sensitive to several latent dimensions, i.e., intervening on some high-correlated latent dimensions will lead to greater semantic concept changes compared to intervene on less important dimensions. Moreover, we also give the APCR distribution information in Figure \ref{fig:APCR}, where the x-axis denote the range-index number (larger index number represents larger APCR values) of APCR values and the y-axis denote the number of dimensions belongs to each range. It furthermore demonstrated that the amount of high-correlated latent dimensions for each concept is small.
\subsection{Find Controlling Set of Latent Dimensions via Optimization}
\begin{figure}[h]
	\centering
	\includegraphics[width=0.45\textwidth]{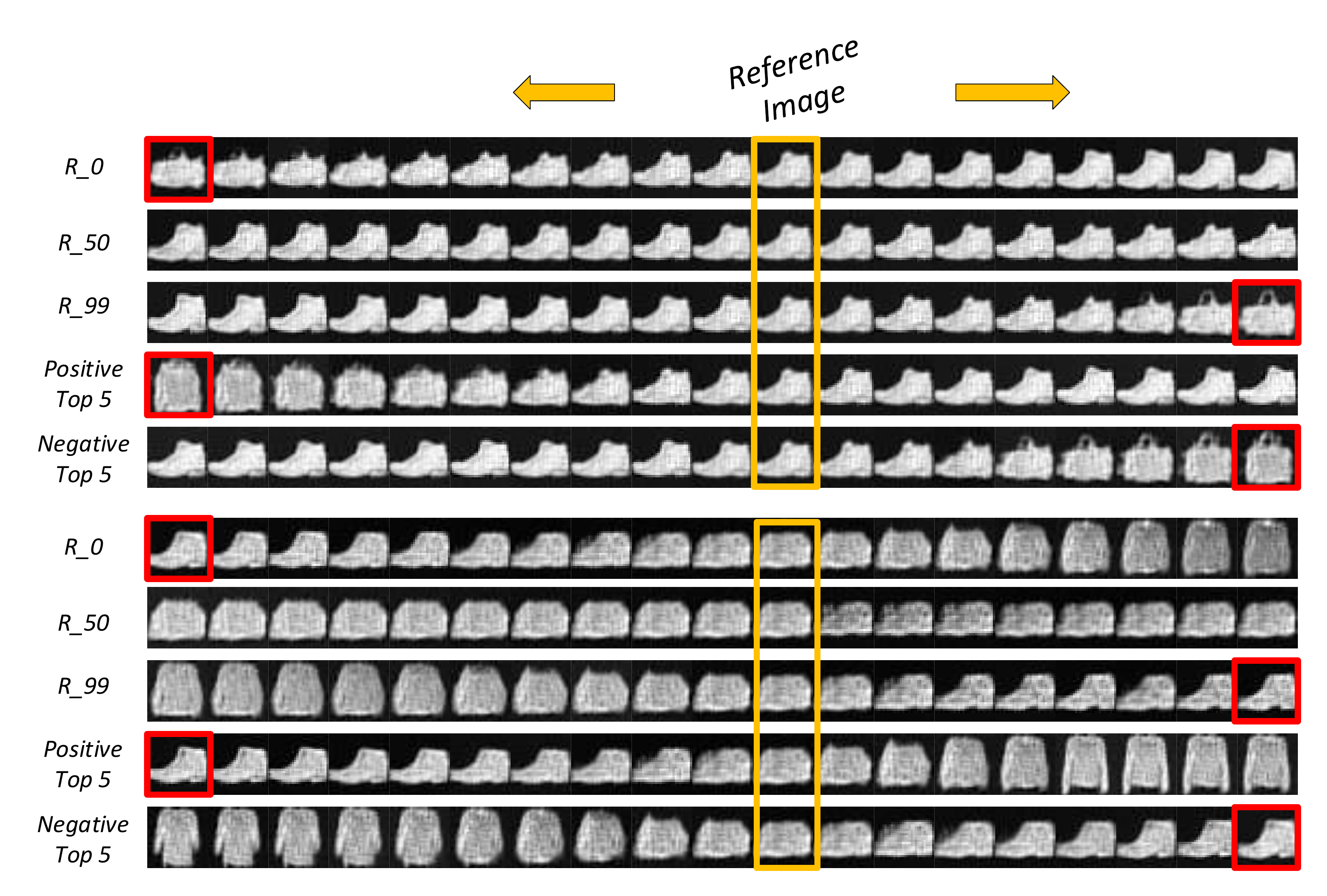}
	\caption{Intervene on controlling set of latent dimensions.}
	\label{fig:ControllingSet}
\end{figure}

We use the optimization-based method described in Section 3.2 to derive the coefficients vector ($w\in[-1,1]$) on latent dimensions for each concept, then we impose thresholds on the coefficients to filter out a positive and negative controlling set of latent dimensions. Larger positive values and smaller negative values denote higher correlation when intervening along the positive and negative direction respectively. We randomly select two classes to demonstrate the controlling latent dimensions, which can be seen in Figure \ref{fig:ControllingSet} (top and bottom parts represent intervention results of class 8 and class9 respectively). As illustrated in Figure \ref{fig:ControllingSet}, the first three rows represent the intervention results of $R_0$,$R_{50}$,$R_{99}$ dimensions respectively, where the numbers denoted as the ranked order, thus $R_0$ and $R_{99}$ represent most correlated latent dimension along bi-directional intervention. The bottom two rows give the intervention results of top-5 positive/ negative correlated dimensions. From the top three rows, we can see that positive intervention results of $R_0$ are similar to negative intervention results of $R_{99}$ while intervene on $R_{50}$ car hardly leads to any generation changes. Moreover, we intervene in top-5 positive and negative dimensions and get better results, which can be seen from the bottom two rows. This indicates that for a given specific concept, we can derive the controlling set of latent dimensions of it, hence concept manipulation can be achieved by intervening on these dimensions (furthermore results can be seen in the next subsection).

We also propose to use intersection ration ($IR_{ctrl}$) as an evaluation metric of the concordance of the controlling dimensions derived by sequential intervention and optimization. The results can be seen in Table \ref*{tab:IntersectionDim}, it indicates that controlling dimensions derived by two methods were quite similar with an average IR score of 0.75. It furthermore demonstrated that latent dimensions contribute differently to specific concept generation.
\begin{table}[h]
	\centering
	\begin{tabular}{c|ccccc}
		\toprule
		Classes &  class0  &  class1  &  class2  &  class3  &  class4\\
		\midrule
		$IR_{ctrl}$&  0.7  &  0.9  &  0.7  &  0.8  &  0.4\\
		\midrule
		Classes&  class5  &  class6  &  class7  &  class8  &  class9\\
		\midrule
		$IR_{ctrl}$&  1  &  0.7  &  0.9  &  0.7  &  0.7\\
		\bottomrule
	\end{tabular}
	\caption{Intersection ration of high-correlated latent dimensions derived by sequential intervention and optimization}
	\label{tab:IntersectionDim}
\end{table}
\subsection{Controllable Concept Manipulation with Controlling Latent Dimensions}
\begin{figure}[h]
	\centering
	\includegraphics[width=0.45\textwidth]{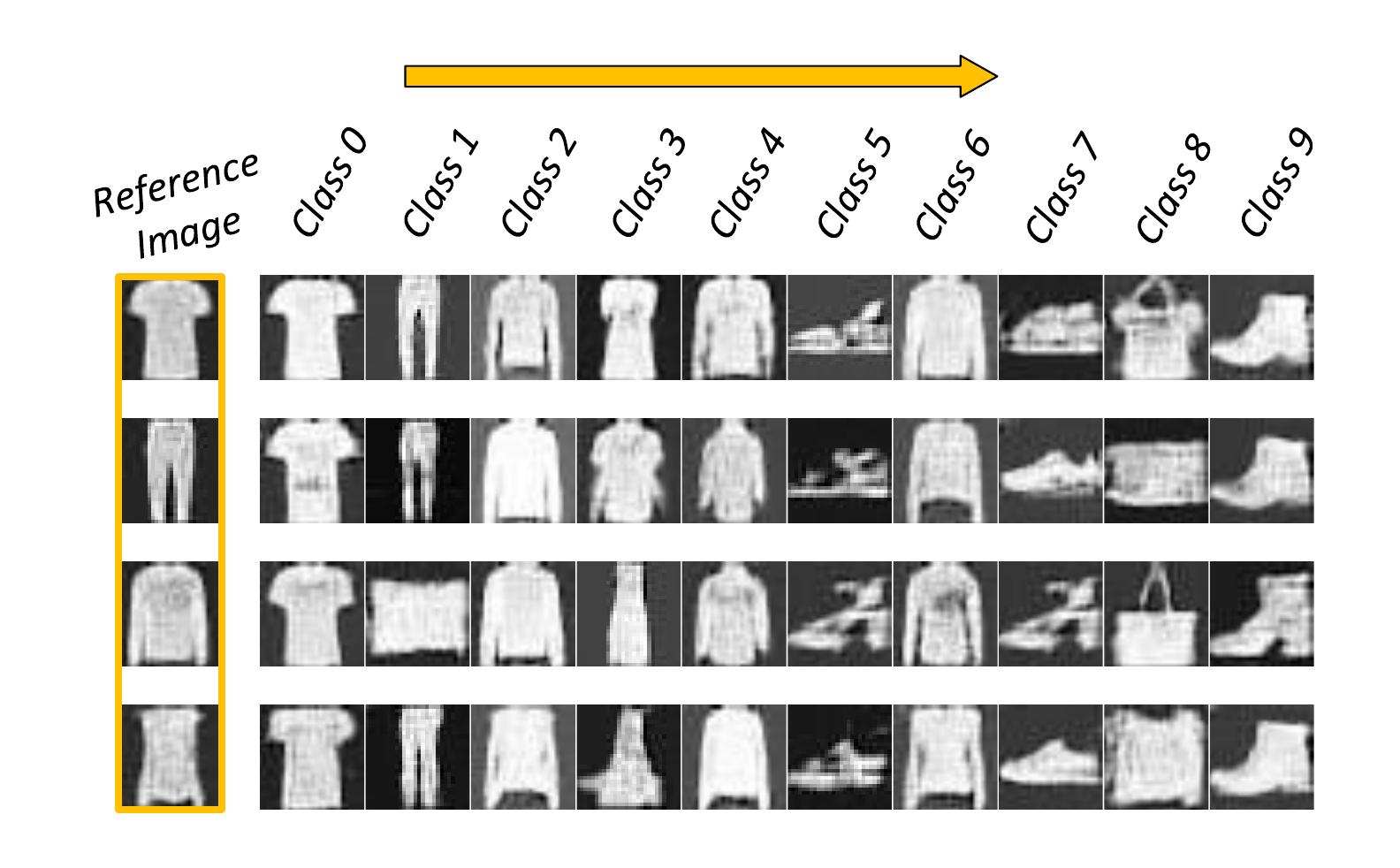}
	\caption{Controllable concept manipulation on Fashion-MNIST through intervening on controlling set of latent dimensions (final manipulation results).}
	\label{fig:Ctrl_Fashion}
\end{figure}

\begin{figure}[h]
	\includegraphics[width=0.45\textwidth]{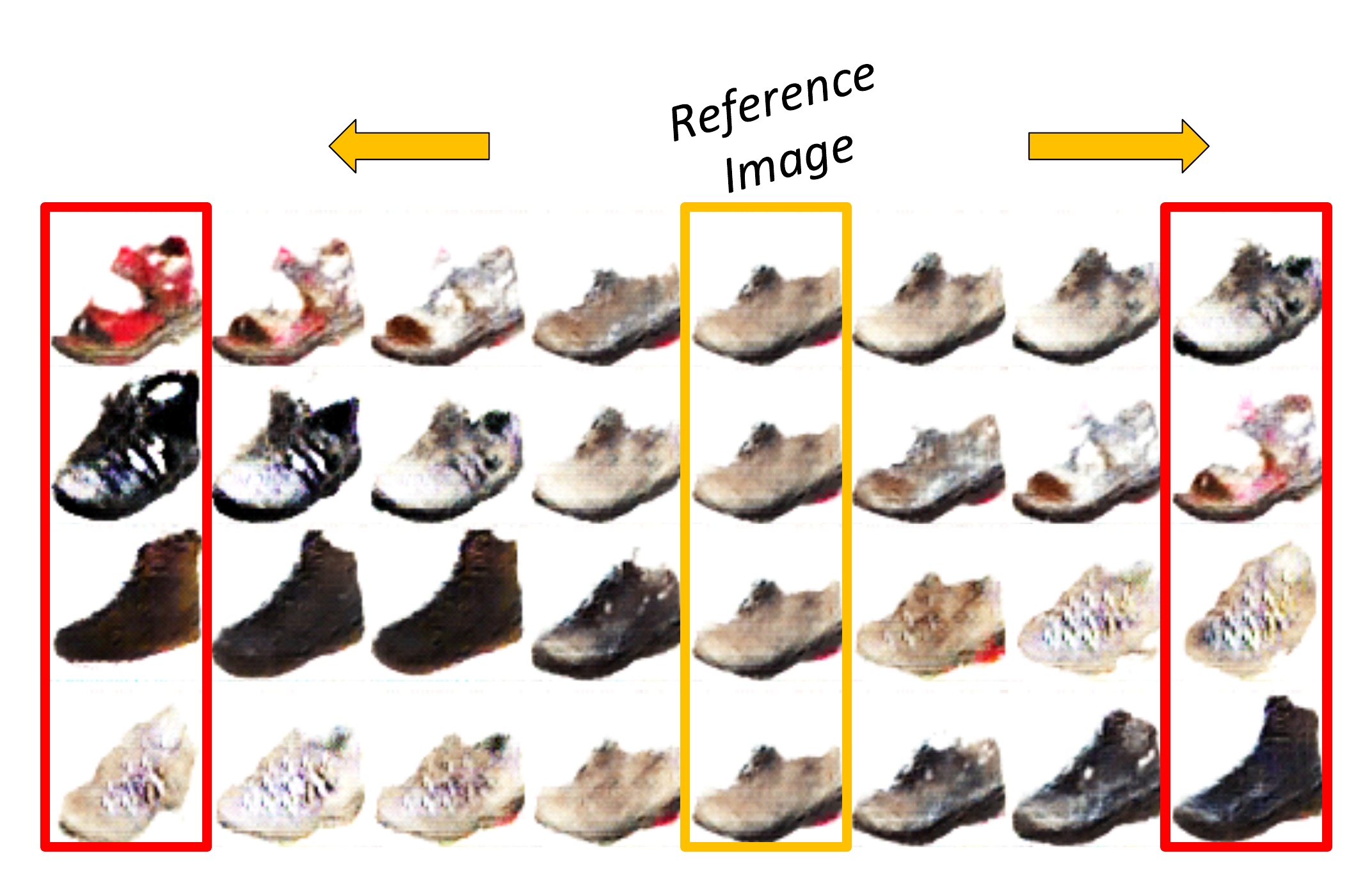}
	\caption{Controllable concept manipulation on UT Zappos50k through intervening on controlling set of latent dimensions.}
	\label{fig:Ctrl_UT}
\end{figure}

As introduced in the above section, we can achieve controllable concept manipulation by intervening the controlling set of latent dimensions. The controllable manipulation results of Fashion-MNIST and UT Zappos50k can be seen in Figure \ref{fig:Ctrl_Fashion} and Figure \ref{fig:Ctrl_UT} respectively. In Figure \ref{fig:Ctrl_Fashion}, the leftmost column denotes the reference images come from different classes of Fashion-MNIST, and the right 10 column denote the corresponding directional manipulation results, i.e. translate the reference image to different classes. Similarly, in Figure \ref{fig:Ctrl_UT}, the middle column in the yellow bounding-box denotes the reference image in UT Zappos50k and other columns denote bi-directional manipulation results. From the results, we can see that controllable concept manipulation can be achieved by intervention on controlling set of latent dimensions.
\subsection{Interesting Findigns of Extreme Intervention on Latent Dimensions}
\begin{figure}[h]
	\includegraphics[width=0.45\textwidth]{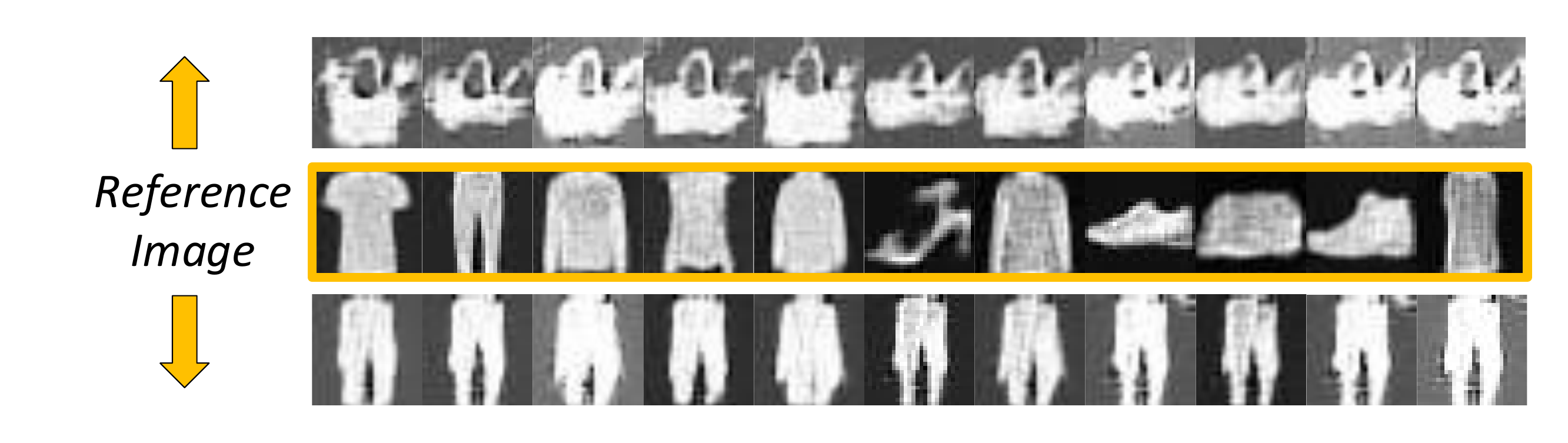}
	\caption{Extreme intervention results by adding impulse stimuli on specific latent dimensions bidirectionally.}
	\label{fig:extreme}
\end{figure}

\begin{figure}[h]
	\includegraphics[width=0.45\textwidth]{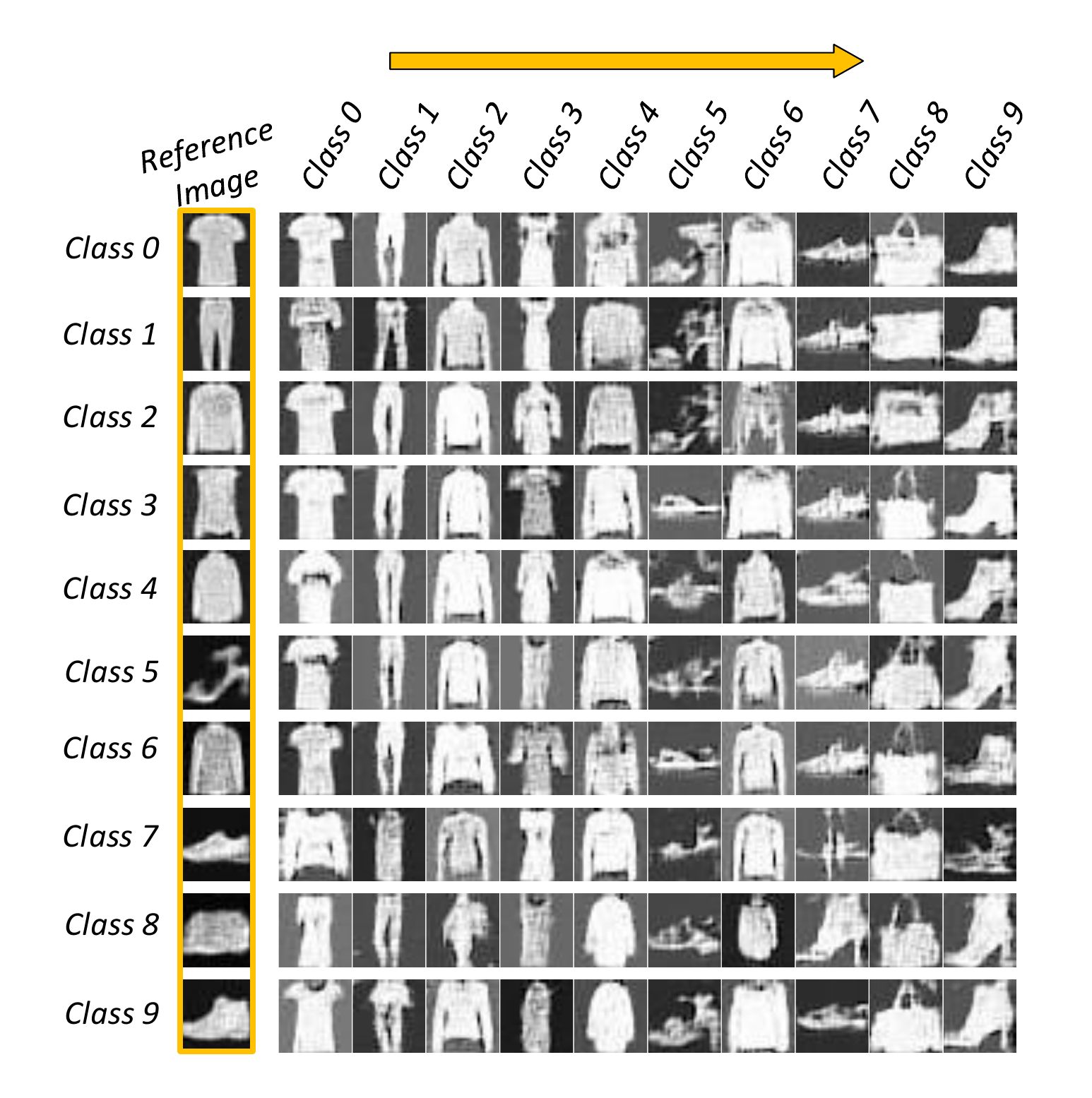}
	\caption{Controllable Class-to-Class translation by extreme intervening on some specific latent dimensions (final translation results).}
	\label{fig:class2class}
\end{figure}

We also have some interesting findings of the latent interventions, which are demonstrated in Figure \ref{fig:extreme}. Usually, the common choice of latent variables is drawn from a Gaussian normal distribution, by now, researchers have not explored extreme cases of latent values. We propose to add positive and negative impulse stimuli on latent dimensions and observe the corresponding semantic content change in the generated image. The results are demonstrated in Figure \ref{fig:extreme}, where the middle row represents the original images come from different classes, the top and bottom row denote positive/ negative extreme intervention results respectively. From the results, we can see that for images from different classes, if we extremely intervene in some latent dimension (adding impulse stimuli), the final manipulation results will converge to some specific classes. Inspired by this finding, we can find these unique dimensions for various class pairs and achieve controllable class-to-class translation (Figure \ref{fig:class2class}).
\section{Conclusion}

In this paper, we proposed a method to interpret the latent space of GANs with correlation analysis for controllable concept manipulation. We utilized a pre-trained classifier to evaluate the concept changes in the generation variations. We first illustrated that the importance of different latent dimensions that contribute to specific concept generation varies greatly. Moreover, we also proposed two methods to find the controlling set of latent dimensions for different concepts, with which we can fulfill controllable concept manipulation. These interesting findings provide a new route for controllable generation and image editing, moreover, it also gives us some thoughtful insight into the interpretability of GANs' latent space.

\section*{Acknowledgments}
This work was partially supported by the National Natural Science Foundation of China under grand No.U19B2044 and No.61836011.

\ifCLASSOPTIONcaptionsoff
\newpage
\fi
\bibliographystyle{IEEEtran}
\bibliography{bare_conf}

\end{document}